\documentclass[10pt,twocolumn,letterpaper]{article}
\usepackage[pagenumbers]{cvpr}
\definecolor{imdomcolor}{HTML}{ED783A}
\definecolor{grdomcolor}{HTML}{86FF33}
\definecolor{txdomcolor}{HTML}{1F77B4}

\newcommand\blfootnote[1]{%
  \begingroup
  \renewcommand\thefootnote{}\footnote{#1}%
  \addtocounter{footnote}{-1}%
  \endgroup
}

\definecolor{cvprblue}{rgb}{0.21,0.49,0.74}
\usepackage[pagebackref,breaklinks,colorlinks,allcolors=cvprblue]{hyperref}
\usepackage[most]{tcolorbox}
\usepackage{marvosym}

\def\paperID{*****} 
\def\confName{CVPR}
\def\confYear{2025}

\title{LayoutGKN: Graph Similarity Learning of Floor Plans}

\author{
Casper van Engelenburg, 
Jan van Gemert, 
Seyran Khademi\\
Delft University of Technology
}

\begin{document}
\twocolumn[{%
\renewcommand\twocolumn[1][]{#1}%
\maketitle
\begin{center}
    \centering
    \captionsetup{type=figure}
    \includegraphics[width=0.9\textwidth]{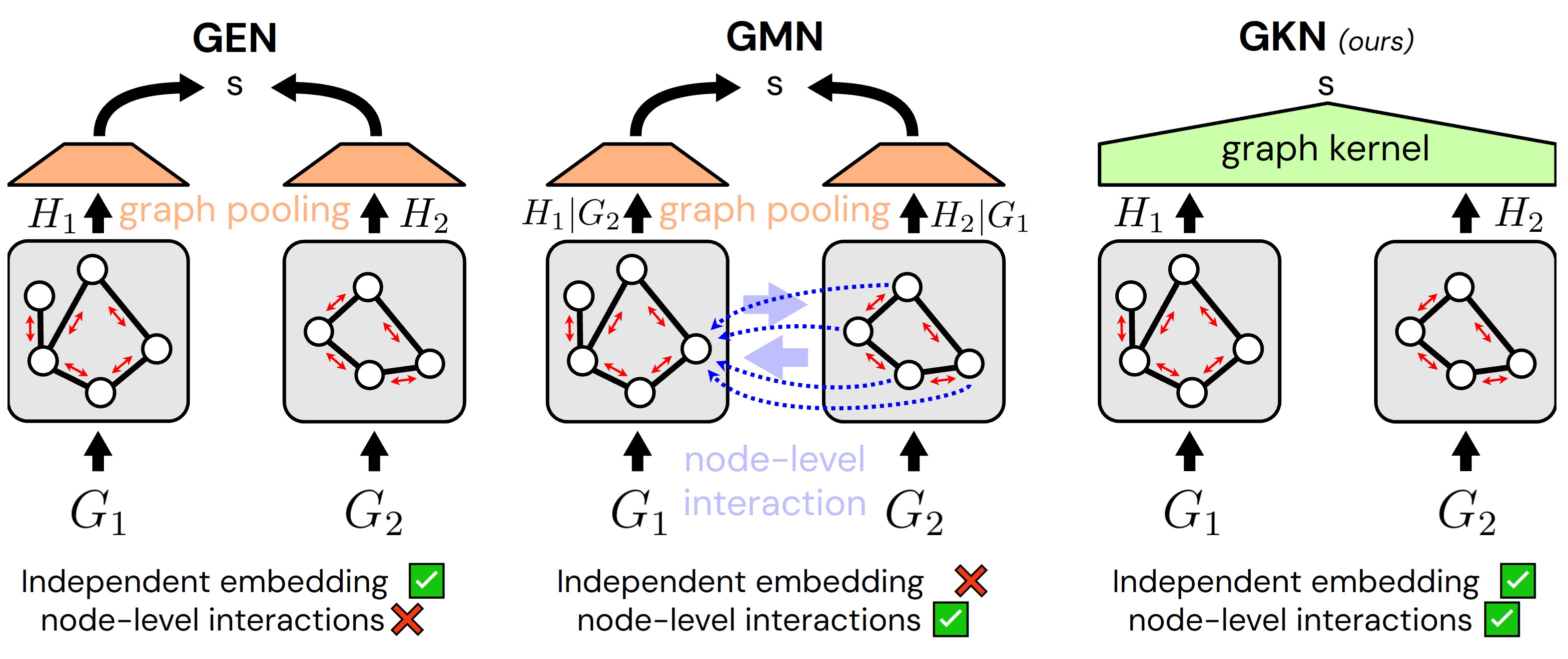}
    \captionof{figure}{ 
        \small 
        \textbf{Joint embeddings architectures for graph similarity learning.}
        (\textbf{\textit{Left}})
        Graph embeddings networks (GENs) independently embed graphs $G_1$ and $G_2$ in a vector space using parameter-shared graph networks (gray box) followed by pooling (orange box).
        The similarity $s$ is computed based on the vector embeddings.  
        (\textbf{\textit{Middle}})
        In addition to GENs, graph matching networks (GMNs) model cross-graph node-level interactions (blue arrows) between the two sets of node features in each layer.
        Embeddings ($H_i$) depend on the other.
        (\textbf{\textit{Right}})
        Our graph kernel network (GKN) shifts node-level interactions to the similarity function via a graph kernel (green box), enabling independent embeddings while preserving node-level interactions.  
        }
    \label{fig:motivational_figure}
\end{center}%
}]
\begin{abstract}
Floor plans depict building layouts and are often represented as graphs to capture the underlying spatial relationships.
Comparison of these graphs is critical for applications like search, clustering, and data visualization. 
The most successful methods to compare graphs \ie, graph matching networks, rely on costly intermediate cross-graph node-level interactions, therefore being slow in inference time. 
We introduce \textbf{LayoutGKN}, a more efficient approach that postpones the cross-graph node-level interactions to the end of the joint embedding architecture.
We do so by using a differentiable graph kernel as a distance function on the final learned node-level embeddings.
We show that LayoutGKN computes similarity comparably or better than graph matching networks while significantly increasing the speed.
\href{https://github.com/caspervanengelenburg/LayoutGKN}{Code and data} are open.
\blfootnote{
To appear in BMVC '25.
Mail: c.c.j.vanengelenburg@tudelft.nl
}
\end{abstract}
\section{Introduction}\label{sec:intro}

In contrast to the natural environment, the built environment follows a remarkably simple spatial logic -- which, more remarkably so, seems to be consistent anywhere on earth.
When stripped of its details, our built environment can be adequately described as a hierarchy of networks: the main infrastructure as a network of roads connecting cities and towns; cities as a network of roads that divide and bring about the composition of neighborhoods and buildings; buildings as stacks of floor plans; floor plans as networks of rooms and corridors; etc.
Often modeled as graphs of nodes and edges, these networks are embedded in physical space, with nodes and edges carrying information about location, shape, and spatial relations.
Encoding such relations in machine-readable form has enabled computer-aided methods to understand~\cite{patil_layoutgmn_2021}, analyze~\cite{pizarro_automatic_2022, hetang_segment_2024}, and synthesize~\cite{nauata_house-gan_2020, he_coho_2024} our built environment.
In this work, we focus on floor plans and floor plan representation learning. 
Nonetheless, most ideas could prove worthy in the other `levels' of our built environment.

As visual depictions of building layouts, floor plans offer an elementary yet powerful abstraction of spatial structure—an aspect central to the quality of space~\cite{alexander_pattern_1977, hillier_space_1976, steadman_architectural_1983}.
Of key importance to floor plan machine understanding is the ability to effectively compare floor plans based on their spatial structure, as well as the ability to do so \textit{quickly} 
-- especially for search-related tasks, data visualization, and clustering or classification.
The question of spatial similarity is inherently subjective, multi-faceted, and task-dependent.
However, we believe, as do others~\cite{patil_layoutgmn_2021, jin_shrag_2022, park_floor_2023, sabri_semantic_2017}, that floor plan similarity can be adequately modeled as a graph comparison problem, given the availability of the graph through image vectorization. 

Most notably, graph matching networks (GMNs)~\cite{li_graph_2019} have been successful at `solving' graph comparison problems, and have been adopted for use in floor plan similarity~\cite{patil_layoutgmn_2021, jin_shrag_2022}.
A GMN is a form of a joint embedding architecture (JEAs), in which two parameter-sharing neural networks (\ie, encoders) process two distinct data samples simultaneously to arrive at two corresponding embeddings~\cite{bromley_signature_1993, chen_simple_2020}.
The similarity is measured based on a distance (\eg, Euclidean) between the embeddings.
For GMNs, the encoders are graph neural networks (GNNs).
The key distinction between classical JEAs is that, in GMNs, information is exchanged between nodes across both graphs during both training and inference.
These cross-graph node-level interactions allow for the explicit modeling of node-level correspondences, which are otherwise thrown away when the graph-level embeddings are computed independently.
Although effective, cross-graph node-level interactions do not allow for the offline computation of embeddings for a given database of floorplans, due to the pairwise dependence of the embeddings. 
Cross-interaction modules hinder the usability of GMNs in real-time retrieval systems where similarity, i.e., the distance between the embeddings, needs to be computed fast.  
Our contributions enable an effective yet efficient solution for graph-based floor plan retrieval, and are summarized as follows: 

\begin{itemize}
    \item We propose LayoutGKN: a GNN model built on a differentiable graph similarity metric, which effectively captures the spatial similarity in floor plan representation learning.
    \item We model the node-level interactions at the end of the JEA by using a differentiable path-based graph kernel.
    \item Experiments on RPLAN and zero-shot generalization to MSD demonstrate that LayoutGKN achieves comparable or better ranking performance than LayoutGMN while being significantly faster. 
\end{itemize}

\section{Related works}\label{sec:related}

\paragraph{Floor plan similarity}
Floor plan retrieval, cast as a problem of similarity learning, in which the goal is to rank a gallery database of floor plans to a query, has been the subject to previous works~\cite{patil_layoutgmn_2021, jin_shrag_2022, park_floor_2023, khade_interactive_2023, takada_similar_2018, sharma_high-level_2018, sabri_semantic_2017, wessel_room_2008, van_engelenburg_ssig_2023}.
Relevant other applications include the retrieval of interface layouts and documents~\cite{vedaldi_learning_2020, patil_layoutgmn_2021, bai_layout_2023, wu_cross-domain_2022} and room layouts~\cite{fisher_characterizing_2011}.
Determining the similarity between two floor plans, as is framed in~\cite{patil_layoutgmn_2021}, is a multi-faceted task, regarding semantic (\eg room functions), geometric (\eg, room shapes), and relational (\eg, permeability or adjacency) information layers.
The \textit{intersection-over-union} (IoU) is often considered as an instrumental measure for spatial similarity~\cite{vedaldi_learning_2020, patil_layoutgmn_2021, jin_shrag_2022}.
While relatively quick to compute, IoU is sensitive to translations, rotations, and scale changes -- and, most importantly, does not explicitly measure the relational similarity between the components that compose the layout~\cite{van_engelenburg_ssig_2023}.
Similar to most other works~\cite{vedaldi_learning_2020, jin_shrag_2022, patil_layoutgmn_2021, khade_interactive_2023, takada_similar_2018, van_engelenburg_ssig_2023, sabri_semantic_2017, park_floor_2023}, we frame floor plan similarity as a graph comparison problem, in which we model floor plans as spatially-attributed graphs of nodes that represent rooms and edges that define their spatial relations.
Unlike previous learning-based methods~\cite{vedaldi_learning_2020, patil_layoutgmn_2021, jin_shrag_2022} that train and evaluate under the IoU metric, we do so using a graph-based similarity metric, which we show better aligns with human judgment of floorplan similarity. 

\paragraph{Graph similarity}
To measure similarity between graphs, many works resort to the use of a \textit{graph edit distance} (GED)~\cite{sanfeliu_distance_1983}, which for example is used to assess compatibility in the generation of floor plans~\cite{nauata_house-gan_2020}, or the \textit{maximum common subgraph} (MCS) used directly for floor plan retrieval~\cite{takada_similar_2018}.
GED is a known NP-complete problem (details in~\cite{zeng_comparing_2009}), exponential in the number of nodes, thus hindering the use where similarity needs to be computed fast (\eg, in search engines).
To address limitations in efficiency, graph kernels (GKs), such as those defined in~\cite{shervashidze_weisfeiler-lehman_2011, feragen_scalable_2013}, have been proposed to solve graph comparison or related problems.
We encourage the reader to read a recent review on GKs ~\cite{kriege_survey_2020}.
While graph kernels are effective for comparing relational structures, a key limitation is their reliance on handcrafted input features, such as node attributes. 
We use graph kernels to compute similarity between floor plans; however, unlike traditional methods, our approach learns the node features directly from data. 

\paragraph{Graph similarity learning}
In turn, learning-based solutions using \textit{graph neural networks} (GNNs) have been proposed for the problem of comparing graphs~\cite{bai_simgnn_2019, ma_deep_2020, li_graph_2019, riba_learning_2020}. 
Most notable is the line on \textit{graph matching networks} (GMNs)\cite{li_graph_2019} that explicitly encodes cross-graph node-level interactions in the joint embedding architecture, which captures fine-grained structural correspondences between the nodes.
Benchmarks on floor plan similarity, such as \textit{LayoutGMN}~\cite{patil_layoutgmn_2021} extensively use GMNs.
Problematic to the use of GMNs is the fact that the node/graph embeddings cannot be computed in isolation~\cite{zheng_grasp_2024}, drastically decreasing efficiency in real-time retrieval.
Instead of modeling expensive cross-graph node-level interactions \textit{across the GNNs}, we only model such interactions at the end of the pipeline using a differentiable path-based graph kernel, \textit{GraphHopper}~\cite{feragen_scalable_2013}, on top of the final node embeddings.
Because the node embeddings can be precomputed independently, our method is much faster than is the case for GMNs (Fig.~\ref{fig:motivational_figure}).
\section{Problem formulation}\label{sec:sim}

\subsection{Floor plans as attributed graphs}
To explicitly model spatial relations, a floor plan is represented as an undirected graph $G = (\mathcal{V}, \mathcal{E})$ of nodes $u \in \mathcal{V}$ that describe rooms connected by edges $(u,v) \in \mathcal{E} \subseteq \mathcal{V} \times \mathcal{V}$ that describe the permeability between the rooms.
We should not forget that the use of the graph as representational device of a floor plan is by no means a new idea, and stems, most notably, from a vast line of works in \textit{Space Syntax} (\eg, as in~\cite{hillier_space_1976} in the mid 70s) and other works around the same time (\eg, in Alexander's~\cite{alexander_pattern_1977} and Steadman's~\cite{steadman_architectural_1983} seminal works).
In these works, only the graph's connectivity (\ie, its topology) was usually explored.
Similar to other works of late~\cite{patil_layoutgmn_2021, nauata_house-gan_2020}, we expand the graphs by populating the nodes and edges with a rich set of attributes, which we do as follows.
Specifically, each node $u$ is endowed with a package of node features. 
1) A one-hot encoding of the category of the room's function $\mathbf{c}^{(u)} \in \mathbb{R}^8$.
Categories are: "living room", "bedroom", "kitchen", "bathroom", "dining", "store room", "balcony", "corridor".
For example, $[1;0;0;0;0;0;0]$ describes the living room.
2) A vector that characterizes the room's shape $\mathbf{s}^{(u)} \in \mathbb{R}^6$, defined as:
$\mathbf{s} = [c_x; c_y; w; h; \sqrt{a}; p/4]$
where $(c_x, c_y)$ denotes the center of the room, $l_x$ the maximum size in $x$ and $l_y$ in $y$, $a$ the area, and $p$ the perimeter. 
3) To make use of the graph kernel,  we need for each node the shortest-path histogram matrix $\mathbf{M}^{(u)}$~\cite{feragen_scalable_2013}.
A path is a sequence of non-repeated nodes connected through edges present in a graph, which for a floor plan captures how one could walk from one space to another.
The shortest path between two nodes is the one which traverses the least edges.
The entry $[\mathbf{M}^{(u)}]_{ij}$ counts how often $u$ occurs at the $i$-th position on shortest paths of length $j$.
We set the size of $\mathbf{M}^{(u)}$ to $\mathbb{R}^{\delta \times \delta}$, in which $\delta$ is the maximum path length, usually taken as the longest shortest path i.e., the graph diameter.
Since graph diameters of floor plan graphs are relatively small, we set $\delta=4$.
Each edge $(u,v)$ carries a vector $\mathbf{e}^{(u,v)}$ indicating the permeability of two adjacent spaces: $[1;0]$ for access connectivity and $[0;1]$ for adjacent-only. 

\subsection{Floor plan similarity as graph comparison problem}
We formulate the challenge of floor plan similarity as a graph comparison problem:
We seek a function $s$, operating on two floor plans represented as attributed graphs 
$G_i$ and $G_j$, $ s(G_i, G_j): \mathcal{G} \times \mathcal{G} \rightarrow \mathbb{R}^+$,
such that, when $G_i$ is similar to $G_j$, $s$ is large; and when dissimilar, $s$ is small.
We narrow down the set of possible solutions by framing the problem in terms of a distance metric learning formulation~\cite{yang_distance_nodate}: 
We seek a learnable function, e.g., an encoder, $f_\theta$, parameterized by $\theta$, that embeds $G$ in a representation space $\mathcal{H}$ (\ie, $H = f_\theta (G)$).  
A similarity measure $s_\mathcal{H}$ is required that computes the similarity between two such representations:

\begin{equation}\label{eq:dml_sim}
    s(G_i, G_j) = s_\mathcal{H} \left( H_i = f_\theta (G_i), H_j = f_\theta(G_j) \right).
\end{equation}

Given the similarity function $s_\mathcal{H}$,  the objective is to learn $f_\theta(G)$ through a differentiable loss.
To decide upon an proper metric for training and evaluation,  a user study is conducted, in which we asked architects to rank floor plans. 
Albeit some level of disagreement, we found a positive and significant correlation between human judgment and a similarity based on a graph edit distance (GED)~\cite{sanfeliu_distance_1983}, while almost none for IoU (→ suppl. mat. for details).
Node and edge categories are considered when computing GED. 
Node categories correspond to the room types and edge categories correspond to the type of room-to-room permeability, which can either be access (\ie, door) or adjacent-only (\ie, wall but no door). 
We consider the weight of each edit operation (\eg, changing node category, deleting / adding edge) the same.
Unlike other learning-based frameworks that train and measure the goodness of retrieval based on IoU~\cite{patil_layoutgmn_2021, jin_shrag_2022, vedaldi_learning_2020}, we train and evaluate based on a normalized variant of GED to express the similarity: 
$\textrm{sGED}(G_i, G_j) = \exp (- \textrm{GED}(G_i, G_j) / (\left| G_i \right| + \left| G_j \right|))$, 
where $\left| G \right|$ is the size of the graph (\ie, the number of nodes).
\section{Method}\label{sec:method}

\subsection{Graph kernel network}
\begin{figure*}[t]
    \centering
    \includegraphics[width=1\textwidth]{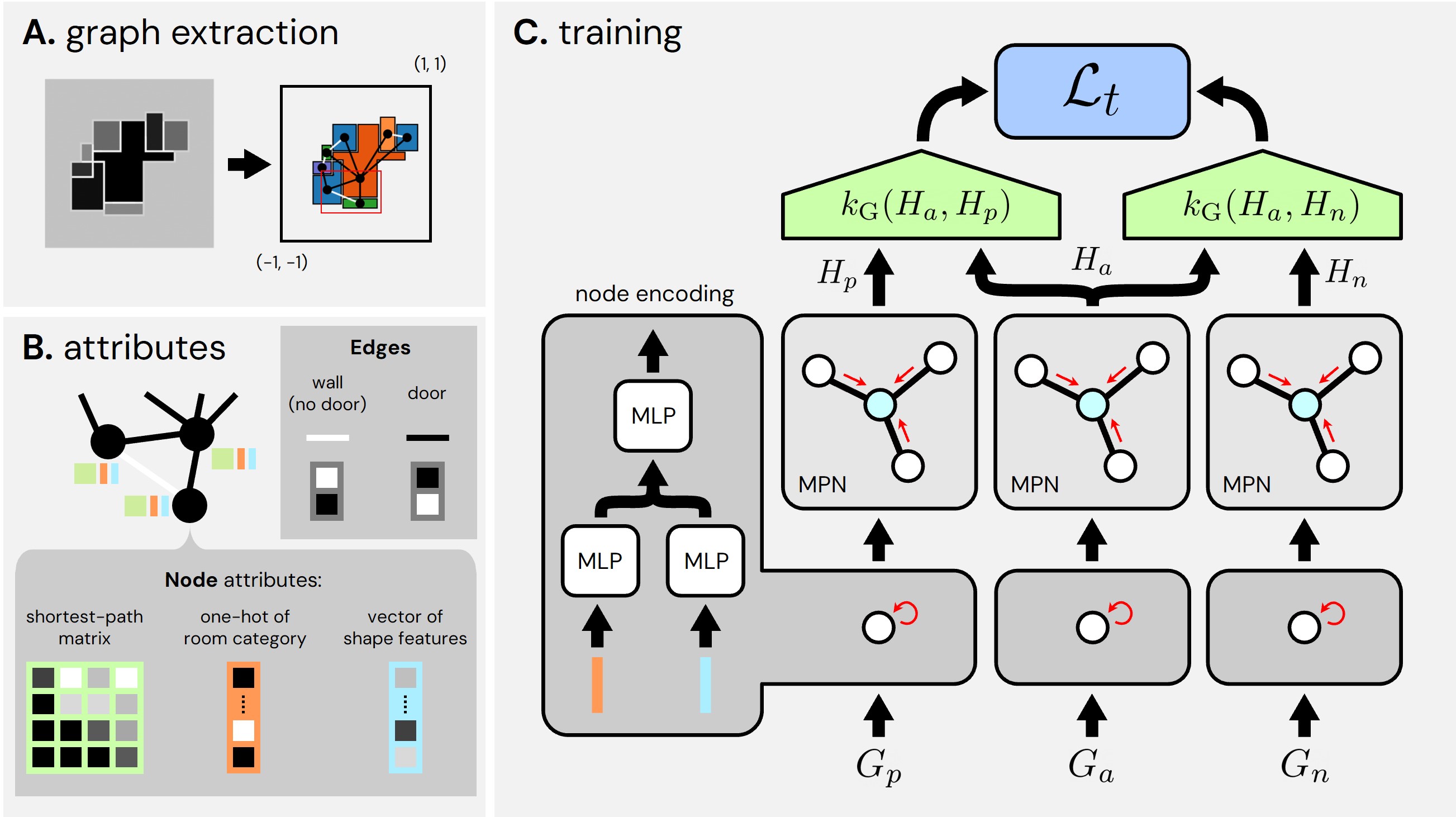}
    \caption{ 
        \small 
        \textbf{Overview of LayoutGKN.}
        (\textit{\textbf{A. graph extraction}}) 
        Semantic images from RPLAN are converted into richly-attributed access graphs. 
        The floor plans' geometries (\ie, the rooms represented as polygons) are centered at $(0,0)$ and scaled to fit within the unit square box.
        The unit box amounts to 20 x 20 meters in reality (\ie, 0.1 equals 1 meter).
        The color indicates the room's semantic category (\eg, dark orange for living room, green for balcony).
        Edges are modeled when two rooms share a door (black) or a wall (white edge).
        (\textit{\textbf{B. attributes}}) 
        Each node represents a room and is endowed with 3 attributes: the shortest-path matrix, a one-hot encoding of the room's category and a vector of shape features.
        (\textit{\textbf{C. training}})
        LayoutGKN is trained using triplets of graphs, each containing an anchor ($G_a$), positive ($G_a$), and negative ($G_a$) graph.
        The goal is to penalize the relative distance between anchor-positive and anchor-negative.
        Anchor, positive, and negative are simultaneously fed into parameter-shared graph neural networks, which consist of a node encoder (dark gray box), followed by a series of $L$ graph message passing network layers (light gray box).
        This results in embedded graphs ($H_i$).
        The anchor-positive and anchor-negative similarities are computed as $s_{\text{ap}} = k_G (H_a, H_p)$ and $s_{\text{an}} = k_G (H_a, H_n)$, respectively.
        The triplet loss $\mathcal{L}_t$ penalizes the relative distance $\log(s_{\text{an}} / s_{\text{ap}})$, given a margin $m$.
    }
    \label{fig:method}
\end{figure*}

The joint embedding architecture of our network contains two elements: 
a graph neural network (GNN) to learn node embeddings, and a graph kernel to compute the similarity between two sets of node embeddings. 
If $f_\theta$ denotes the GNN and $s_\mathcal{H}$ the similarity function, our framework, coined as graph kernel network (GKN), computes the similarity between graphs as given in Eq~\ref{eq:dml_sim}.
How are methods fits the others is highlighted in Fig.~\ref{fig:motivational_figure}.
An overview of our method is given in Fig.~\ref{fig:method}.

\paragraph{Graph neural network for feature learning}
Node feature vectors are encoded using separate MLPs: $\mathbf{f}_{\text{cat}}: \mathbb{R}^8 \rightarrow \mathbb{R}^d$ on top of the room function vector; $\mathbf{f}_{\text{shape}}: \mathbb{R}^6 \rightarrow \mathbb{R}^d$ on the room shape vector; and $\mathbf{f}_{\text{edge}}: \mathbb{R}^2 \rightarrow \mathbb{R}^d$ on the edge vector.
The outputs of $\mathbf{f}_{\text{cat}}$ and $\mathbf{f}_{\text{shape}}$ are concatenated and subsequently fed into another MLP, $\mathbf{f}_{\text{node}}: \mathbb{R}^{2d} \rightarrow \mathbb{R}^d$, to form the first hidden state:

\begin{equation}
    \mathbf{h}_0^{(u)} = 
    \mathbf{f}_{\text{node}} 
    \left(
        \begin{bmatrix}
            \mathbf{f}_{\text{cat}} (\mathbf{c}^{(u)})
            \\
            \mathbf{f}_{\text{shape}} (\mathbf{s}^{(u)})
        \end{bmatrix}
    \right).
\end{equation}

To learn expressive node embeddings, we use a message passing network (MPN) similar to the one in~\cite{li_graph_2019}.
The MPN is a product of $L$ consecutive graph convolutional layers. 
Each layer contains a message and an update function. 
The message to node $u$ in layer $l$, $\mathbf{m}^{(\rightarrow u)}_{l}$, aggregates node information from its direct neighbors $v\in\text{ne}(u)$ (and itself) and takes into account the type of edges between them.
Specifically, for each neighbor $v$, the hidden node features of the node itself $\mathbf{h}_l^{(u)}$, the one of the neighbor $\mathbf{h}_l^{(v)}$, and the edge encoding $\mathbf{r}^{(u,v)}$ are concatenated and fed into a shared MLP $\mathbf{f}_{\text{intra}}:\mathbb{R}^{3d} \rightarrow \mathbb{R}^{d}$. 
We use average aggregation over all the individual messages and arrive at the following definition for $\mathbf{m}^{(\rightarrow u)}_{l}$:

\begin{equation}
\mathbf{m}^{(\rightarrow u)}_{l} = 
    \sum_{v\in\text{ne}(u)} 
    \mathbf{f}_{\text{intra}}
    \left(
        \begin{bmatrix}
            \mathbf{h}^{(u)}_{l} \\
            \mathbf{h}^{(v)}_{l} \\
            \mathbf{r}^{(u, v)}
        \end{bmatrix}
    \right).
\end{equation}
\noindent To arrive at the node features at layer $l+1$, the aggregated message and the node feature itself are stacked and fed into a GRU, $\mathbf{f}_{\text{update}}:\mathbb{R}^{2d} \rightarrow \mathbb{R}^{d}$:

\begin{equation}\label{eq:update}
    \mathbf{h}^{(u)}_{l+1} = 
    \mathbf{f}_\text{update} 
    \left(
        \begin{bmatrix}
            \mathbf{h}^{(u)}_{l} \\
            \mathbf{m}^{(\rightarrow u)}_{l}
        \end{bmatrix}
    \right).
\end{equation}

\paragraph{Graph kernel as similarity function}
We use the GraphHopper path-based graph kernel~\cite{feragen_scalable_2013} to compute the similarity between two graphs.
Specifically, given two graphs and their corresponding sets of learned node embeddings by the GNN, the graph kernel similarity $k_G$, as shown by the authors, can be computed in a remarkably simple form: 
as a weighted sum over node kernels across all pairs of nodes and their corresponding vector embeddings:

\begin{equation}\label{eq:ghopper_kernel}
    k_G(G_1, G_2) = 
    \sum_{u \in \mathcal{V}_1} \sum_{v \in \mathcal{V}_2} 
    \left\langle \mathbf{M}^{(u)}, \mathbf{M}^{(v)} \right\rangle  
    \cdot 
    k_{\text{node}} \left( \mathbf{h}_L^{(u)}, \mathbf{h}_L^{(v)} \right),  
\end{equation}

\noindent where the node kernel $k_{\text{node}} \left( \bullet, \bullet \right)$ computes the similarity between two nodes, and is typically Gaussian ($\mu \propto d^{-1}$~\cite{kriege_survey_2020}):

\begin{equation}\label{eq:node_kernel}
    k_{\text{node}}\left( \mathbf{h}_L^{(u)}, \mathbf{h}_L^{(v)} \right) = 
    \exp\left(
        -\mu{\left\| \mathbf{h}_L^{(u)} - \mathbf{h}_L^{(v)} \right\|^2}
    \right).
\end{equation}

\subsection{Training and inference}
We train LayoutGKN under a triplet network setting~\cite{hoffer_deep_2015}; thus, penalizing \textit{relative} distances. 
Given a triplet of graphs $(G_a, G_p, G_n)$, in which $a$, $p$, and $n$ denote anchor, positive, and negative.
We feedforward each graph through $f_\theta$ resulting in embeddings $H_a$, $H_p$, and $H_n$.
We use a normalized variant of the graph kernel (Eq.~\ref{eq:ghopper_kernel}) to compute the similarity between two such embeddings:
$
s_{\mathcal{H}} (H_i, H_j) = 
k_G(H_i, H_j) /
\sqrt{k_G(H_i, H_i) \cdot k_G(H_j, H_j)}.
$
Given that we compute the relative distance from anchor-positive to anchor-negative as $d_{\text{apn}} = s_{\mathcal{H}} (H_a, H_n) / s_{\mathcal{H}} (H_a, H_p)$, we formulate the triplet loss as

\begin{equation}\label{eq:kernel_loss}
\begin{aligned}
    \mathcal{L}_t &= 
    \left[ 
        m + 
        \log 
        \left( 
        d_{\text{apn}}
        \right) 
    \right]_+ \\
    &= 
    \left[ 
        m + 
        \log 
        \left( 
            \frac{k_G (H_a, H_n)}{k_G (H_a, H_p)} \sqrt{\frac{k_G (H_p, H_p)}{k_G (H_n, H_n}} 
        \right) 
    \right]_+,
\end{aligned}
\end{equation}

\noindent where $[\bullet]_+ = \max (\bullet, 0)$. 
We use the log to effectively penalize small relative distances more. 
Because the node (Eq.~\ref{eq:node_kernel}) and graph kernel (Eq.~\ref{eq:ghopper_kernel}) are differentiable w.r.t. to the node embeddings, the loss (Eq.~\ref{eq:kernel_loss}) is as well.
The second line in~\ref{eq:kernel_loss} shows an efficient implementation of our loss in which $k_G(H_a, H_a)$ is canceled out.
Similar to~\cite{patil_layoutgmn_2021, jin_shrag_2022, vedaldi_learning_2020}, we mine triplets based on the computable metrics we aim to mimic.
Our objective, as mentioned before in Sec.~\ref{sec:sim}, is different though: we mimic sGED not MIoU. 
To find informative triplets, we first rank each floor plan in the training set on MIoU, filter the best 50, and re-rank them on sGED.
Each triplet is formed by the query and some combination of positive and negative found in the re-ranked list.
Given some anchor, we first pick a positive s.t. $0.9 > \text{sGED}(G_a, G_p) > 0.6$.
To ensure hard negatives, we pick a negative s.t. $0.7 < \text{sGED}G_a, G_n) / \text{sGED}(G_a, G_p)< 0.9$.
In the case of GKN, during training, we pre-compute the shortest-path histogram matrices. During inference, we pre-compute the final node embeddings and denominator of $s_\mathcal{H}$ as well.
\section{Results and discussion}
\label{sec:results}

\subsection{Experiment setup}

(\textbf{\textit{Datasets}})
We train and test on RPLAN~\cite{wu_data-driven_2019} and evaluate generalization to MSD~\cite{van_engelenburg_msd_2024}. 
After careful cleaning of both datasets, RPLAN and MSD contain 46K+ and 16K+ apartment-level floor plans, respectively.
Statistics, descriptions, and pre-processing steps can be found in the suppl. mat. We train our models on RPLAN, and split training and test data with a ratio of 8:2.
(\textbf{\textit{Baselines}})
We compare LayoutGKN, which we will refer to as GKN in the remainder of the text, with \textit{LayoutGMN}~\cite{patil_layoutgmn_2021}, in short GMN, and basic baseline methods including GEN and GK.
GK is the \textit{GraphHopper} graph kernel as-is, for which we model the node features by concatenating $\mathbf{s}^{(u)}$ and $\mathbf{g}^{(u)}$. 
For all methods, we use the same graph representation, as given in Sec.~\ref{sec:sim}. 
In the case of GEN, GMN, and GKN, the same node encoder and intra-graph message passing mechanisms are used. 
In the case of GMN, an inter-cross message $\mathbf{mc}^{(\rightarrow u)} \in \mathbb{R}^d$ is concatenated to the input of $\mathbf{f}_\text{update}$ (Eq.~\ref{eq:update}), equivalent to~\cite{patil_layoutgmn_2021} (Eq.~4, pp.~4). 
For GMN and GEN, we use the same graph pooling mechanism as in~\cite{li_graph_2019} and use the conventional triplet margin loss on the relative distances between the final graph-level vector embeddings: 
$\mathcal{L}_t = [ d_\text{an} - d_\text{an} + m]_+$. 
(\textbf{\textit{Evaluation}})
For comparing the methods, we report the triplet accuracy, Precision at 5 and 10 (P@5 and 20), and inference times ($t$). 
We use 4-fold cross-validation, and report the average scores across the folds on the test set. 
To evaluate the generalization of the methods, we report the zero-shot precision scores on MSD only.
(\textbf{\textit{Implementation}})
Each model is trained for at most 200 epochs, with early stopping (patience is 10), on an NVIDIA GeForce RTX 4090 Ti GPU using AdamW~\cite{loshchilov_decoupled_2019} with a learning rate of $10^{-4}$ and batch size of 64.
We use layer and batch normalization for the node encoder parts and message passing updates, respectively.

\begin{table*}[ht]
\centering
\caption{
    \textbf{Performance comparisons on RPLAN and MSD.} 
    We report: the triplet accuracy; precision (P) scores at 5 and 10; and inference time $t$ per 10K pairs.
    Best in \textbf{bold}.
    Note that the graph kernel (GK) does not involve any learning: it is GraphHopper kernel~\cite{feragen_scalable_2013} as-is, for which we model the node features by concatenating the categorical and shape-based node features (\ie, for each node $u$: 
    $[\mathbf{c}^{(u)}; \mathbf{s}^{(u)}]$).
    GK is a strong baseline for retrieval.
    During inference, embeddings ($H_i$) can only be precomputed for GEN and LayoutGKN, which creates the order of magnitude difference in inference time between these methods and LayoutGMN.
    * We train and represent the data in LayoutGMN~\cite{patil_layoutgmn_2021} in exactly the same way as is the case for the other methods, which is slightly different from the original implementation in which the triplets are mined using MIoU and floor plans are represented as fully connected graphs.
    } 

\begin{tabular}{l|cccc|cc}
    \toprule
    & \multicolumn{4}{c|}{\textbf{RPLAN}} & \multicolumn{2}{c}{\textbf{Zero-shot to MSD}} \\
    & Accuracy ({\small $\uparrow$}) 
    & P@5 ({\small $\uparrow$}) & P@10 ({\small $\uparrow$}) 
    & $t$ [s] ({\small $\downarrow$}) 
    & P@5 ({\small $\uparrow$}) & P@10 ({\small $\uparrow$})\\
    \midrule
    
    LayoutGK (\textit{baseline}) & 65.63{\footnotesize $\pm$0.00}
    & 0.389{\footnotesize$\pm$0.000}
    & 0.439{\footnotesize$\pm$0.000}
    & 1.2{\footnotesize$\pm$0.4} 
    & na
    & na
    \\
    
    LayoutGEN (\textit{baseline}) 
    & 96.24{\footnotesize $\pm$0.07} 
    & 0.603{\footnotesize $\pm$0.007} 
    & 0.665{\footnotesize $\pm$0.004}
    & \textbf{0.7{\footnotesize $\pm$0.1}}
    & 0.595{\footnotesize $\pm$0.015} 
    & 0.605{\footnotesize $\pm$0.018} 
    \\

    LayoutGMN* {\footnotesize \textcolor{blue}{CVPR'21}}
    & \textbf{97.74{\footnotesize $\pm$0.05}} 
    & 0.616{\footnotesize $\pm$0.004} 
    & 0.675{\footnotesize $\pm$0.002} 
    & 35.6{\footnotesize $\pm$10.5}
    & 0.585{\footnotesize $\pm$0.026} 
    & 0.596{\footnotesize $\pm$0.020} 
    \\

    LayoutGKN (\textit{ours}) 
    & \textbf{97.78{\footnotesize $\pm$0.10}}
    & \textbf{0.623{\footnotesize $\pm$0.004}} 
    & \textbf{0.683{\footnotesize $\pm$0.002}} 
    & 1.8{\footnotesize $\pm$0.5} 
    & \textbf{0.674{\footnotesize $\pm$0.024}} 
    & \textbf{0.697{\footnotesize $\pm$0.017}} 
    \\
    
    \bottomrule
\end{tabular}
\label{tab:main_results}
\end{table*}

\begin{figure*}[ht]
    \centering
    \includegraphics[width=0.8\textwidth]{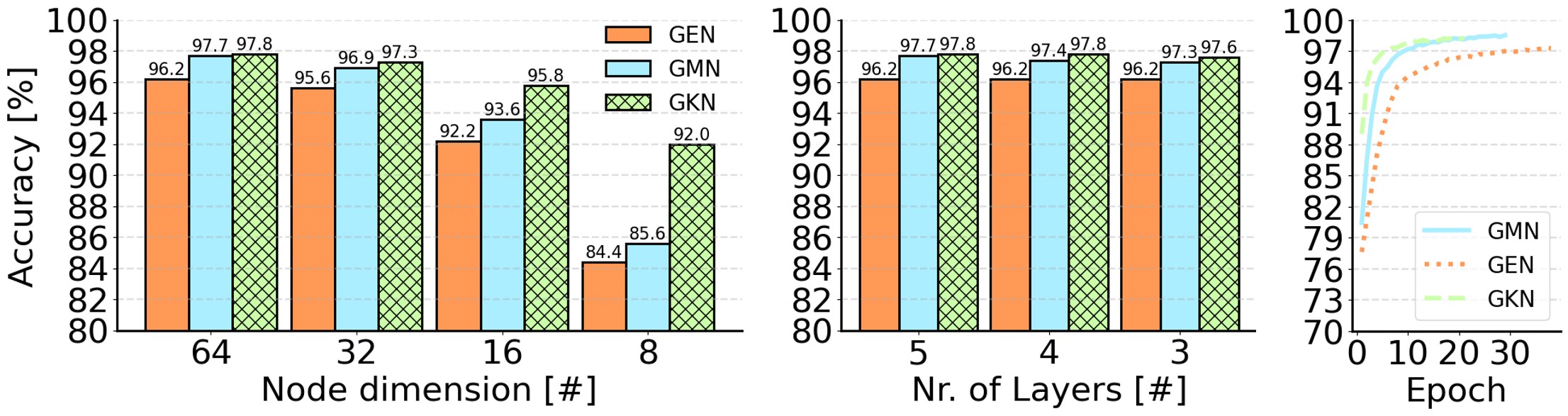}
    \caption{ 
        \small 
        (\textbf{\textit{Left and middle}}) 
        The effect of the number of learnable parameters on the triplet accuracy.
        We vary the number of parameters by changing the hidden node dimension and number of graph convolutional layers.
        The hidden node dimension has a profound effect on the accuracy: the smaller the node dimension, the more a decrease in accuracy.
        The rate at which the accuracy drops for decreasing node dimension is lowest in the case of GKN.
        In the case of the number of layers there are negligible differences in accuracy (across all methods).
        (\textbf{\textit{Right}}) 
        Typical training curves (on the validation split) showing that: 1) GKN converges the fastest and 2) has a head-start because of the kernel.
    }
\label{fig:size}
\end{figure*}

\subsection{Effectiveness}
Table~\ref{tab:main_results} reports, on the left side of the vertical line, the overall effectiveness of all methods. 
Although similar to GMN in accuracy, GKN outperforms all other methods on ranking. 
The difference between GKN (or GMN) and GEN scores shows that explicit inclusion of cross-graph node-level interactions leads to a significant increase in accuracy and precision.
However, we show that modeling expensive cross-graph node-level interactions \textit{across the GNNs} might not be necessary for effective graph similarity computation.
Instead, which is the case in GKN, the node-level interactions can be "postponed" to the end, when the final node embeddings are already computed, without losing quality.

\paragraph{Effect of size}
Fig.~\ref{fig:size} shows the effect of network size on triplet accuracy, varying the hidden node dimension ($d$) and number of graph convolutional layers ($L$). 
When changing $d$, we fix $L=5$; when changing $L$, we fix $d=64$. 
We observe that performance drops much faster for GEN and GMN than for GKN as $d$ decreases. 
We attribute this to the different roles of the embeddings.
In GEN and GMN, node embeddings must be sufficiently large to capture both (i) their topological role in the graph and (ii) their semantic attributes (e.g., geometry and room function in floor plans).
Although GMNs explicitly model cross-graph node interactions, they still depend on expressive node embeddings.
In contrast, GKN leverages a topology-aware similarity function (\ie, the graph kernel), which naturally accounts for the topology of the graphs when computing the similarity between them.
As a result, node embeddings in GKN can be smaller without too much of a performance drop, since they need not encode topology as strongly themselves.
We have not yet tested with more `expressive' GNN encoders, such as GAT~\cite{velickovic_graph_2018} or GIN~\cite{xu_how_2019}, but the same capacity limitation seems likely to remain.

\begin{figure*}[ht]
    \centering
    \includegraphics[width=1\textwidth]{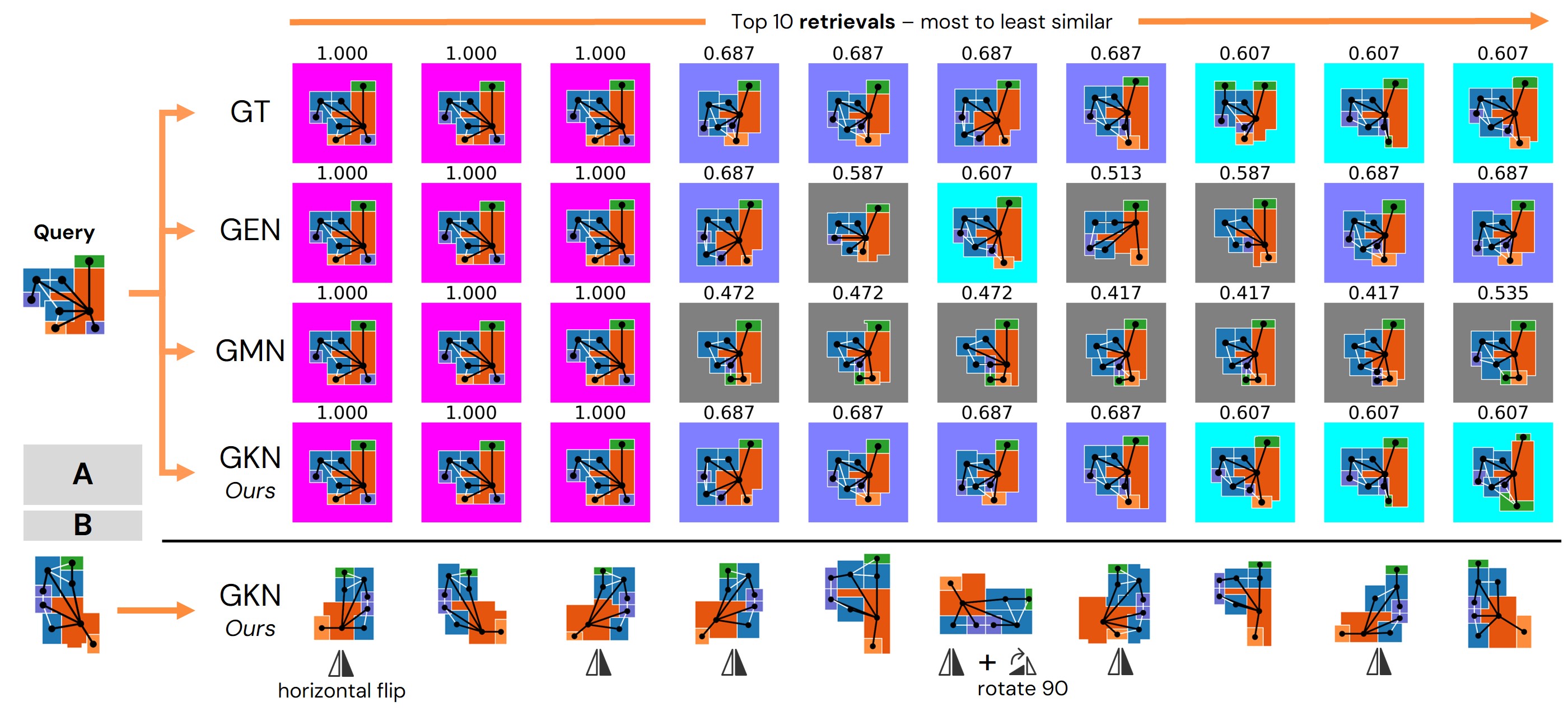}
    \caption{ 
        \small 
        \textbf{Examples of ranking on RPLAN.}
        (\textbf{A})
        Given a query floor plan on the left, the top row shows the ground truth (GT) ranking. 
        The value on top of the floor plan indicates the similarity in terms of sGED. 
        Since the GT scores are quite often equivalent for different retrievals, we color-coded them: the same color means the same GT score. 
        Rows 2 to 4 depict the rankings based on GEN, GMN, and GKN.
        On top of the floor plans, we provide the GT score and indicate using the same color scheme as is the case in the top row where a floor plan would have landed in the GT ranking. 
        A gray background means that it would fall outside the top-10 on GT.
        (\textbf{B})
        We show an example of ranking on GKN in which the top-10 retrievals show signs that the retrieval results are invariant to the elementary grid transformation (flips and 90 degree rotations).
    }
    \label{fig:qual}
\end{figure*}

\paragraph{Qualitative studies}
Fig.~\ref{fig:qual} 
(\textbf{A}) shows a case of ranking under the sGED metric on RPLAN. 
A useful property of training floor plan similarity under a sGED metric, is that retrievals are (practically) invariant to grid transformations. 
In Fig.~\ref{fig:qual} (\textbf{B}), the emergence of rotational and flip invariance is depicted.
Because of the two first entrees ($c_x^{(u)}$ and $c_x^{(u)}$) of the room shape vector $\mathbf{s}^{(u)}$, it is important to note that the model is not strictly invariant to such grid transformations. 
Nonetheless, if you want the model to be \textit{provably} invariant to such grid transformations, simply remove the first two entrees of $\mathbf{s}^{(u)}$.

\subsection{Efficiency}
In the case of GMN, we cannot pre-compute the node embeddings, because of the cross-graph node-level interactions. 
Depending on $d$, $L$, and $N_i = \left| G_i\right|$ for $i \in (1,2)$, the number of FLOPs is order of magnitudes larger for GMNs: $\approx$ 5-10k FLOPs in the case of GKN and GEN; $\approx$ 10-20M FLOPs in the case of GMN.
Therefore, the differences in inference time $t$, shown on the direct left of the vertical line in Tab.~\ref{tab:main_results}, are not surprising: GMN takes $\approx$ 20X more time than the rest.
In real-time search engines, speed is of great importance: results should pop-up in seconds, not minutes.
GMNs compute the similarity between 10K pairs in about 30-40 seconds.
GKNs, on the other hand, compute the similarity between 10K pairs in about 1 to 2 seconds: a drastic speedup compared to GMNs -- a speedup that, most importantly, aligns with what we are after in real-time search.

\subsection{Generalization}
We further assess the models in a zero-shot setting (\ie, no additional training or fine-tuning) by evaluating directly on the more complex MSD dataset~\cite{van_engelenburg_msd_2024}. 
As shown on the right side of Tab.~\ref{tab:main_results}, all methods retain relatively high precision despite the distribution shift from RPLAN to MSD. 
For GKN, precision is even higher on MSD than on RPLAN, and it achieves the strongest performance overall. 
A plausible explanation is that MSD, which contains entire building complexes, exhibits a higher proportion of near-duplicate or structurally similar floor plans, thereby boosting retrieval scores. 
The advantage of GKN compared to the other methods may further stem from the use of the graph kernel, which provides a strong structural prior aligned with the GED-based evaluation and is thus less dependent on dataset-specific differences. 
These results indicate stronger zero-shot transfer for GKN, though disentangling intrinsic model robustness from dataset effects remains an open question.
\section{Conclusion}\label{sec:conclusion}

We introduced LayoutGKN, a graph-based joint embedding architecture for measuring the spatial similarity between floor plan graphs. 
Unlike graph matching networks (GMNs), LayoutGKN decouples representation learning from similarity computation: node embeddings are learned independently, while important cross-graph node-level interactions are only imposed in the distance function via a differentiable path-based kernel. 
Our results on floor plan similarity learning indicate that cross-graph node-level interactions are not necessarily required within the graph encoders themselves, but can be effectively postponed to the similarity-based loss: compared to LayoutGMN, our method achieves comparable or better performance on floor plan retrieval on RPLAN, while drastically increasing the speed.
We are curious whether our method generalizes to other graph domains (\eg, molecules) as well.
\paragraph{Acknowledgments}
We thank all participants of our user study for their valuable time and feedback.
{\small\bibliographystyle{ieeenat_fullname}\bibliography{main}}
\appendix
\onecolumn
\section*{Supplementary materials}

\begin{figure*}[ht]
    \centering
    \includegraphics[width=\textwidth]{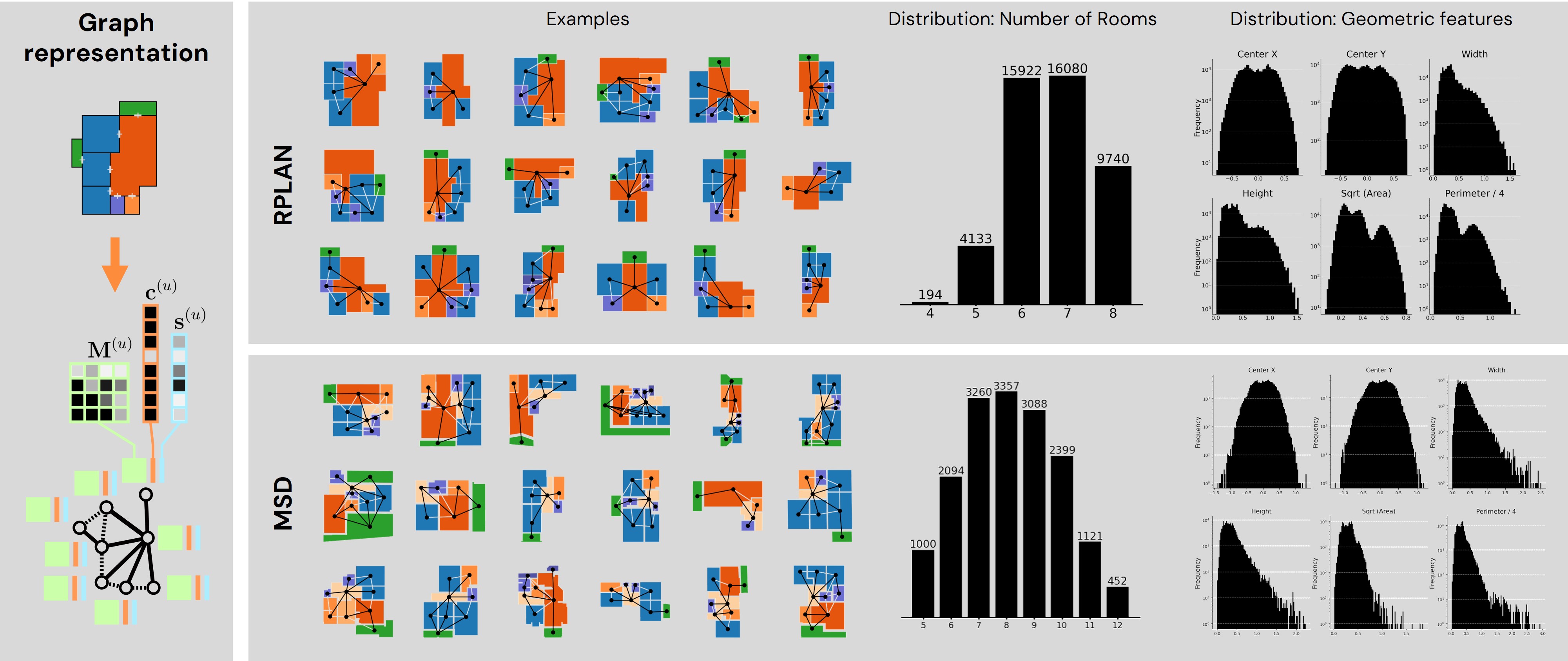}
    \caption{ 
        \small 
        \textbf{Floor plan datasets and representation.}
    }
    \label{fig:datasets}
\end{figure*}

\subsection*{Data}
\paragraph{Datasets: RPLAN and MSD}
We use RPLAN~\cite{wu_data-driven_2019} and MSD~\cite{van_engelenburg_msd_2024}.
RPLAN contains 88K+ floor plans and covers Asian residential apartments. 
As shown by~\cite{van_engelenburg_ssig_2023}, RPLAN contains a substantial amount of near-duplicates as well as floor plans that are \textit{not} entirely connected (\ie, floor plans that contain rooms or room constellations that are disconnected from the rest).
We remove near-duplicates as well as unconnected plans.
The number of remaining floor plans is 46K+.
MSD originates from~\cite{standfest_swiss_2022} and contains 6K+ floor plans of residential building complexes found in Switzerland.
We extract the apartment-level floor plans by cutting the corresponding floor plan graphs along the entrances to the public spaces.
The resulting number of floor plans is 16K+.

\paragraph{Homogenization}
We extract the graphs from RPLAN ourselves and re-model the ones from MSD.
We do so such that the graph formats are homogeneous: floor plans are centered around the origin, the same scale is used (one unit of measurement amounts to 10 meters in reality), and the node and edge features are the same size and contain the same information, in the same order.
Each floor plan is modeled as a graph $G = \left( \mathcal{V}, \mathcal{E} \right)$ of nodes $u \in \mathcal{V}$ that correspond to the rooms and edges $(u,v) \in \mathcal{E}=\mathcal{V} \subseteq \mathcal{V}$ connecting the nodes and that indicate permeability of the rooms.
Each node $u$ is endowed with a package of node features: a category for room's function $c^{(u)} \in \mathbb{N}^8$; a vector that characterizes the room's shape $\mathbf{s}^{(u)} \in \mathbb{R}^6$.
The room shape vector is defined as follows:
$\smash{
\mathbf{s}^{(u)} = 
[
c_x^{(u)},
c_y^{(u)},
w^{(u)},
h^{(u)},
\sqrt{a^{(u)}},
\frac{p^{(u)}}{4}
]^\top}
$.
Here, $(c_x, c_y)$ denotes the center of the room, $l_x$ the maximum size in $x$ and $l_y$ in $y$, $a$ the area, and $p$ the perimeter.
Each edge $(u,v)$ carries a vector $\mathbf{e}^{(u,v)}$ indicating the permeability of two adjacent spaces: $[1;0]$ for access connectivity and $[0;1]$ for adjacent-only.
A visual clarification and some additional statistics are provided in Fig.~\ref{fig:datasets}.

\subsection*{Intersection over union}
We define the \textit{mean intersection over union} (MIoU) between two floor plans $X_i$ and $X_j$ as follows. 
For each room category $c$, let $r_c(X)$ denote the region of $X$ occupied by class $c$ 
(either as a set of pixels in a semantic mask or as the union of all polygons with label $c$). 
We consider the set of categories 
$\mathcal{C}_* = \{\, c \mid \mu(r_c(X_i) \cup r_c(X_j)) > 0 \,\}$, 
i.e.\ all classes that occur in at least one of the two plans, where $\mu(\cdot)$ denotes area 
(pixel count or polygon area). The MIoU is then given by

\begin{equation}\label{eq:miou}
    \mathrm{MIoU}(X_i, X_j) \;=\; 
    \frac{1}{|\mathcal{C}_*|}
    \sum_{c \in \mathcal{C}_*}
    \frac{\mu\!\left(r_c(X_i) \cap r_c(X_j)\right)}
         {\mu\!\left(r_c(X_i) \cup r_c(X_j)\right)}.
\end{equation}

\subsection*{Graph edit distance}
The \textit{graph edit distance} (GED)~\cite{sanfeliu_distance_1983} between two floor plan graphs $G_i$ and $G_j$ is defined as the minimum number of edit operations required to transform $G_i$ into $G_j$, where operations include inserting, deleting, or relabeling nodes and edges (for example, changing a node label from ``living room'' to ``kitchen''). 
We convert GED into a normalized similarity score by scaling with the total number of nodes and applying an exponential decay:

\begin{equation}\label{eq:sged}
    sGED(G_i, G_j) \;=\; 
    \exp\!\left(
        - \frac{GED(G_i, G_j)}{|G_i| + |G_j|}
    \right),
\end{equation}

\noindent where $|G|$ denotes the number of nodes in graph $G$.

\subsection*{Precision}
P@k measures the fraction of relevant items among the top-$k$ retrieved results:  

\begin{equation}
    P@k = \frac{1}{k} \sum_{i=1}^k \mathbf{1}\{\,r_i \text{ is relevant}\,\},
\end{equation}

where $r_i$ is the item at rank $i$ and $\mathbf{1}\{\cdot\}$ is the indicator function.  
In practice, we take the top-50 results returned by the model for each query.  
We then re-rank them according to the ground-truth similarity values.  
Finally, we compute P@k between this ground-truth order and the model’s original ranking.

\subsection*{User study}

\begin{figure*}[ht]
    \centering
    \includegraphics[width=\textwidth]{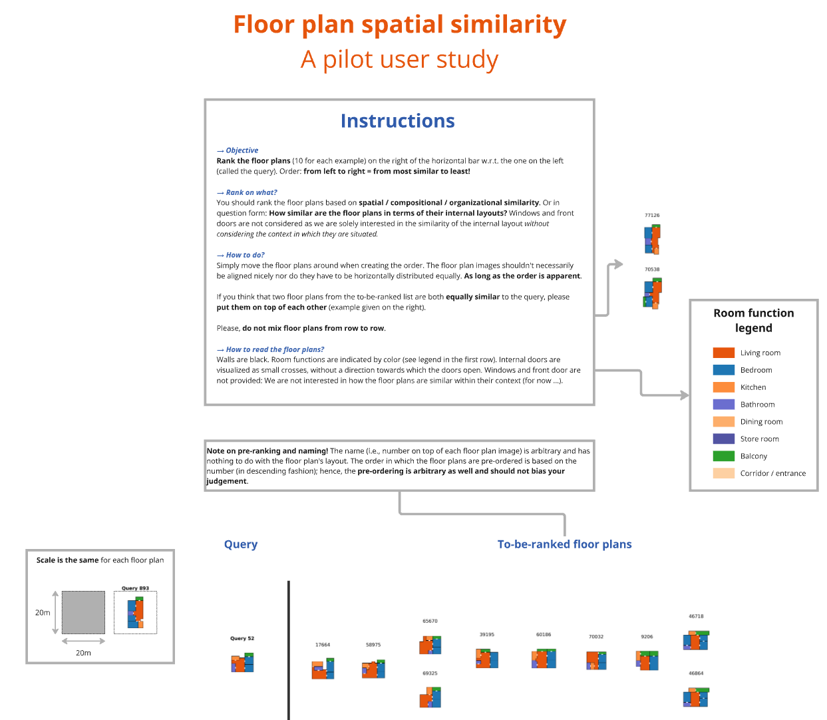}
    \caption{ 
        \small 
        \textbf{Setup of user study in \textit{Miro}.}
    }
    \label{fig:miro}
\end{figure*}

\begin{figure*}[ht]
    \centering
    \includegraphics[width=0.85\textwidth]{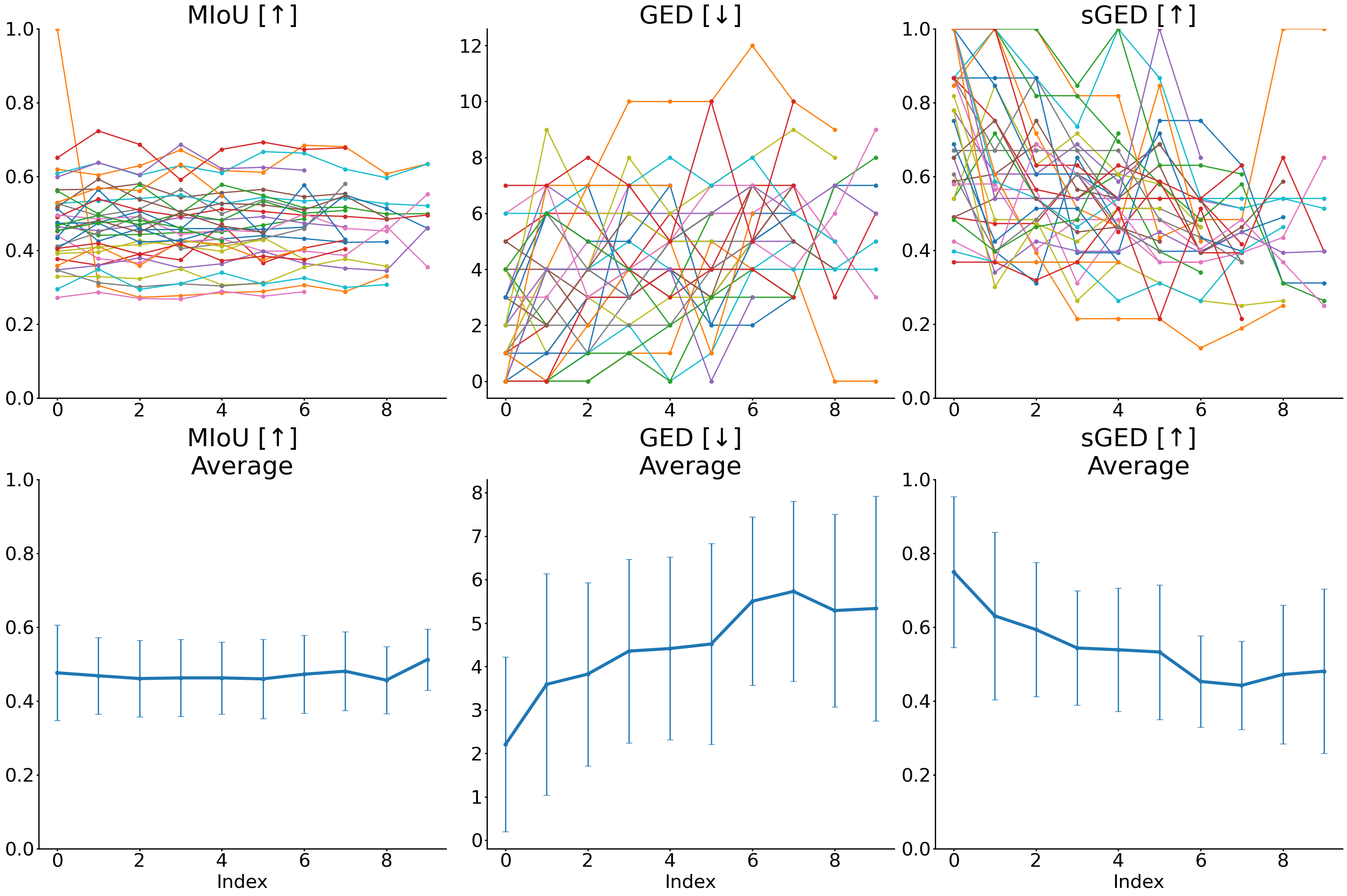}
    \caption{ 
        \small 
        \textbf{User study results.}
    }
    \label{fig:user_results}
\end{figure*}

We conducted a user study to check how trained architects would rank floor plans on spatial similarity. 
We randomly sampled 50 floor plan queries and gathered for each floor plan, the top-10 retrievals on MIoU (Eq.~\ref{eq:miou}).
The top 10 retrievals are randomly shuffled and presented to the architects.
The architects are asked to rank the shuffled floor plans on similarity.
We used \textit{Miro}\footnote{https://miro.com/index/} for getting the annotations: an online software platform which made it possible to easily drag and drop the floor plans.
A snapshot of the setup is given in Fig.~\ref{fig:miro}.
We investigate the correlation between human judgment and the computable metrics (\ie, MIoU and sGED).
In Fig.~\ref{fig:user_results}, we show how the computable metrics evolve for the top-10 floor plans, when ranked by the architects.
The horizontal axis indicates the index of the floor plan where it is placed according to the architects: from most (index = 0) to least (index = 9).
The vertical axis indicates the score of the computable metric.
The first row of plots show the individual curves; the second the mean scores.
Clearly, there is very prominent correlation between human judgment and a similarity based on GED (\ie sGED). 
This means that, in general, architects tend to rank floor plans (in the case that they are already quite similar in terms of overall shape) more on connectivity / topology than on the overlap of spaces.
Therefore, our objective is to mimic sGED in the representation space.

\subsection*{Hyperparameters}
Across all methods, we perform a hyperparameter analysis -- and report the ones with the highest scores.
The hyperparameters we test against are: the learning rate $(1e^{-5} \cdots 1e^{-3})$; the batch size $\{ 8, 16, 32, 64, 128\}$; the node-, edge-, and graph-level hidden dimensions $\{ 8, 16, 32, 64\}$; the number of graph convolutional layers $\{ 2, 3, 4, 5\}$; and the triplet margin $\{ 5e^{-4}, 1e^{-4}, \cdots, 5e^{-1}, 1\}$.

\end{document}